\documentclass{article}
\usepackage{spconf,amsmath,graphicx,booktabs,url}
\usepackage{enumitem}

\def\etf{ET}
\newcommand{\armnone}{\textsc{none}}
\newcommand{\armasr}{\textsc{asr}}
\newcommand{\armaj}{\textsc{asr+json}}
\newcommand{\armturn}{\textsc{turn}}
\newcommand{\armturnasr}{\textsc{turn-asr}}
\newcommand{\armturngold}{\textsc{turn-gold}}
\title{Heard Enough to Act: Early Tool Calling in Spoken Dialogue Systems}
\name{Yizhou Peng$^1$, Chen Chen$^1$, Donghang Wu$^{1,2}$, Hexin Liu$^1$, Eng Siong Chng$^1$}
\address{$^1$ Nanyang Technological University, Singapore \quad $^2$ AI Singapore, Singapore}

\begin{document}
\ninept
\maketitle

\begin{abstract}
Latency-sensitive spoken agents must know not only which tool to call, but also when they have heard enough to act. Waiting for the turn to complete delays execution even when the request is already actionable. However, neither existing spoken agents nor spoken dialogue corpora define the earliest point at which a request becomes actionable. We introduce a synthetic spoken-dialogue pipeline that provides earliest-actionable-point supervision, marking when the observed speech grounds a tool and its required arguments. Through controlled task generation, speech synthesis and alignment, and gated annotation, we construct 26,694 dialogues totaling 786 hours across 48 domains, with disjoint training and test argument values. We additionally create an audio rendering of $\tau^2$-bench for external evaluation. We train an omni-modal language model on this corpus, enabling causal, chunk-level decisions and partial-transcript generation for downstream call completion. The model achieves tool-state macro-F1 scores of 0.48 and 0.47 on held-out dialogues and the external benchmark, respectively. For correct calls, the estimated mean lead over a turn-end cascade, using the same formatter, is 1.8–2.0 seconds in-domain and 1.1 seconds on the $\tau^2$. These results demonstrate a data-driven path toward spoken agents that act while listening. We will open-source the pipeline and the training scripts.
\end{abstract}
\begin{keywords}
full-duplex interaction, low latency, tool calling, spoken dialogue systems
\end{keywords}

\section{Introduction}
\label{sec:intro}

People keep conversation flowing by planning their reply while the other person is still talking, so turns follow within a fraction of a second \cite{stivers,levinson}. A tool-using voice agent, cascaded or end-to-end, still waits for the user to finish before doing anything: only then does it work out what was said, commit to a tool and its arguments, place the call, and await the result; these delays stack rather than overlap. 
Even systems engineered to respond within a couple of hundred milliseconds start that clock at the end of the turn \cite{llamaomni}, and the wait routinely reaches seconds, which users can notice \cite{maslych}. 
We do not attack transcription speed or the tool's own round-trip here; we ask whether \textbf{the step in the middle, deciding on the call and filling in its arguments, must wait for the turn to end at all}.

Much of the field has made this wait \emph{feel} shorter instead. Incremental generators bridge it with fillers, the way people say ``um'' while composing a reply \cite{skantzehjalmarsson}; a recently deployed dialogue robot even detects mid-utterance that a request is becoming clear, yet spends that head start on a spoken preface while the lookup still waits for the turn to end \cite{okafuji}. 
SHANKS \cite{shanks} reasons while listening and completes more than half of its tool calls before the user finishes; Stream RAG \cite{streamrag} post-trains a speech model to issue tool queries during ongoing speech; and DuplexSLA \cite{duplexsla} decodes the user's audio, the agent's speech, and a dedicated planning-and-tool-call stream on one shared clock for car-cabin and smart-home schemas. 
All three learn their timing from proxies, LLM-judged offsets, or change-detection labels; none defines the moment at which a call first becomes decidable, verifies that its labels sit at that moment, or scores timing against it.

The rest of the landscape leaves that moment equally untouched. 
Full-duplex models such as Moshi and SyncLLM \cite{moshi,syncllm} and chunk-level turn-taking predictors \cite{freezeomni,minmo,neuralfsm,vap} decide chunk by chunk who holds the floor; the newest think while the user speaks in explicit tokens \cite{chronological} or latent embeddings \cite{flair}, yet the choice remains silence, thought, or speech, never a tool call.
Incremental dialogue research has shown that meaning is often settled before an utterance ends \cite{iumodel,devault}, but has never tied that insight to an irreversible external action.
And tool-calling benchmarks score which call an agent produces from a completed turn \cite{taubench,tau2bench,bfcl}; even their recent extensions into audio and full-duplex voice, including a rendering of $\tau^2$-bench itself, test whether speech changes \emph{what} to call or whether the task still completes, not \emph{when} the call became callable \cite{texttovoice,audio2tool,tauvoice}. 
In short, no existing system fixes \textbf{the earliest instant} at which the tool and every argument are grounded in the audio, or scores its timing against that instant by construction.

In this paper, we aim to fire the tool call at the moment the user has said enough, and doing so is as much a data problem as a modeling one: no spoken task-oriented corpus, human-recorded \cite{spokenwoz} or LLM-generated \cite{sdfactory}, marks the earliest instant at which a call's tool and all its arguments are grounded in the audio heard so far, and generated corpora invite the train--test contamination that inflates scores \cite{contamination}.  
We built a synthetic spoken-dialogue pipeline that stamps every call with that instant, keeping training and test argument values disjoint by construction, and adapted an omni-modal LLM with LoRA to judge whether the instant has arrived, then either emit the call itself or hand off a short transcribed span for downstream completion.
On held-out data and on our own audio rendering of $\tau^2$-bench \cite{tau2bench}, our model, when it fires at the right moment, recovers the correct call more often than not and starts it an estimated 1.8--2.0\,s early on average; the remaining bottleneck is the firing decision itself, which misses a fair share of call opportunities and triggers on some turns it should leave alone.
Our contributions are: 
\begin{itemize}[nosep,leftmargin=*]
\item a 786\,h synthetic spoken-dialogue corpus with gated, alignment-anchored earliest-actionable-point label pipeline and a leakage-free split by construction;
\item a single-stream control-token model that decides, per chunk, whether that point has arrived;
\item a payload study on what the model should emit when it fires, so that an early fire becomes a correct tool call.
\end{itemize}

\begin{figure*}[t]
  \includegraphics[width=0.95\textwidth]{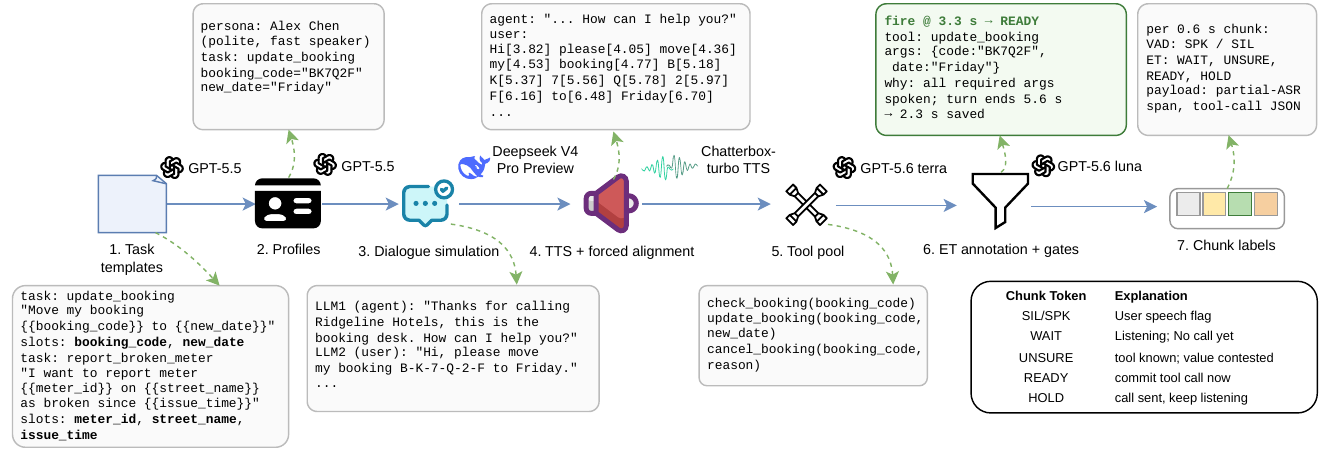}
  \vspace{-0.8em}
  \caption{Data pipeline. Typed task templates are instantiated by a deterministic entity sampler, simulated as text dialogues, rendered to aligned speech, and annotated with gated early-tool-call events; the green event carries the earliest-actionable-point label ($t_{\text{fire}}$) and its saved time. The table on the right glosses the per-chunk labels.}
  \label{fig:pipeline}
  \vspace{-1.5em}
\end{figure*}

\section{Methods: Data Pipeline and Modeling}
\label{sec:methods}

We cast early tool calling (\etf{}) as a per-chunk decision problem. User audio arrives in 0.6\,s chunks; after each chunk, the model emits a voice-activity token and a tool state in \{\textsc{wait}, \textsc{unsure}, \textsc{ready}, \textsc{hold}\} on speech, as shown in Fig.~\ref{fig:pipeline}.
\textsc{ready} commits mid-utterance to a specific call and \textsc{unsure} flags a likely but under-specified tool; either emission is a \emph{fire}. Each gold call is an event $(\text{tool}, \text{args}, t_{\text{fire}})$, where $t_{\text{fire}}$ is the earliest instant at which every required argument is grounded in the speech heard so far, and $\Delta_{\text{save}} = t_{\text{end}} - t_{\text{fire}}$ is its \emph{gold saving} over a turn-end call.

\subsection{Data pipeline}
\label{ssec:data}

Figure~\ref{fig:pipeline} shows the generation stack. 
GPT-5.5 authors 48 domains with 18--19 task \emph{templates} each (900 tasks) whose identity fields are typed entity slots; a deterministic, collision-checked sampler instantiates $30 \times 900 = 27{,}000$ profiles (306 yield no usable dialogue, leaving 26{,}694); two DeepSeek-V4Pro Preview sessions \cite{deepseekv4} simulate the user and agent scripts; Chatterbox-turbo \cite{chatterbox} synthesizes each turn with per-persona cloned voices; and the NeMo forced aligner \cite{nfa} provides word-level timestamps.

Tool-call supervision is annotate-then-gate.
Each profile receives a synthesized tool pool (mean 4.5 tools), GPT-5.6-Terra proposes candidate events, and three deterministic gates verify every proposal: Gate~1 discards non-executable calls (unknown tool, malformed or type-mismatched arguments); Gate~2 collects arguments that the transcript does not ground or that a later utterance supersedes; Gate~3 resolves them, repairs the argument, or delays $t_{\text{fire}}$ to the instant the argument is actually grounded.
The funnel prunes $140{,}595 \to 78{,}193 \to 53{,}492$ events, and with $t_{\text{fire}}$ anchored to the forced alignment, $\Delta_{\text{save}}$ has mean 2.40\,s, median 1.36\,s, and p90 6.60\,s. Since the gates can only delay, $t_{\text{fire}}$ is an upper bound on the true earliest point.
We then expand events onto the 0.6\,s grid: every chunk carries a VAD label (\textsc{spk}/\textsc{sil}); the speech chunk at $t_{\text{fire}}$ takes \textsc{ready} or \textsc{unsure}, and the other speech chunks read \textsc{wait} before the turn's first fire and \textsc{hold} after it, an anti-duplicate guard that a fresh \textsc{ready} (a correction or a second tool) can still break.
Table~\ref{tab:data} summarizes the corpus. Splitting by \emph{task-group holdout} leaves zero train/test argument-value overlap. 
For open benchmarking, we ran the identical pipeline on $\tau^2$-bench's own tasks and tool schemas \cite{tau2bench}. Unlike $\tau$-Voice \cite{tauvoice}, this rendering scores the \emph{firing decision} per chunk, not task completion.

\begin{figure*}[t]
  \centerline{\includegraphics[width=\textwidth]{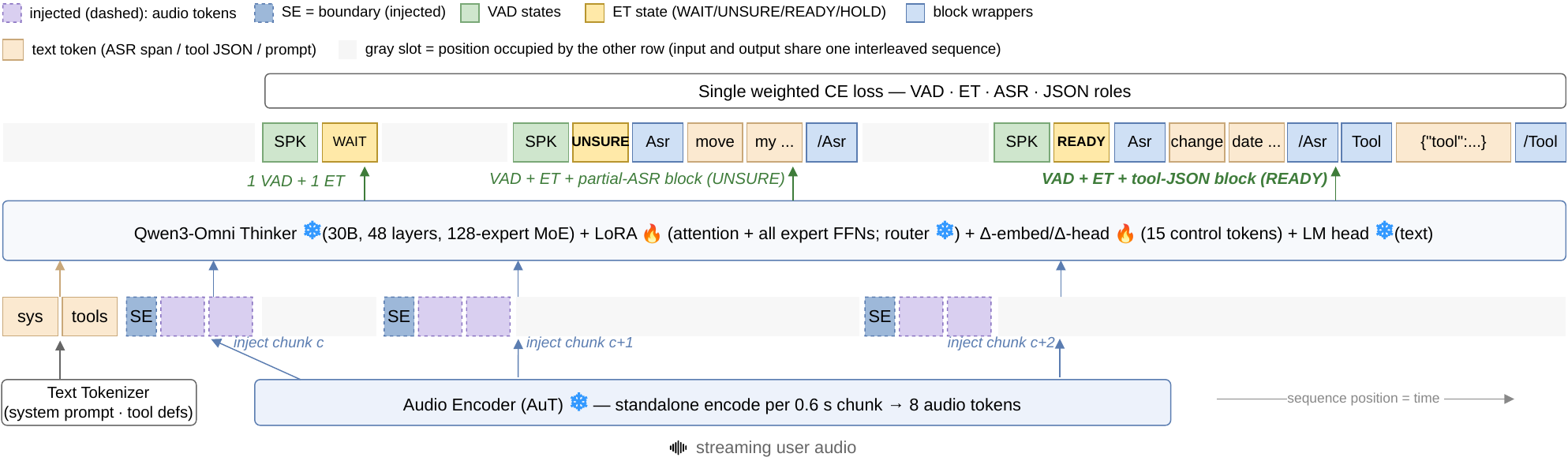}}
  \vspace{-1.0em}
  \caption{Single-stream model design. One autoregressive, KV-cached sequence interleaves injected audio chunks with emitted control tokens and payload blocks; one LM head decodes VAD, \etf{}, ASR, and tool JSON under a single weighted cross-entropy loss. Snowflakes are frozen; flames are trainable. The example shows the \armaj{} variant of Eq.~\eqref{eq:variants}.}
  \label{fig:model}
\vspace{-1.0em}
\end{figure*}

\subsection{Modeling}
\label{ssec:model}
Fig.~\ref{fig:model} shows the model. We keep Qwen3-Omni-30B-A3B \cite{qwen3omni,qwen3} frozen, add fifteen control tokens (VAD and \etf{} states, payload delimiters, pending markers) whose embedding and output rows are initialized to the mean of related words and trained only through small \emph{delta} vectors, as in prompt tuning \cite{prompttuning}, and attach LoRA \cite{lora} to attention and all per-expert FFN projections.

\begin{table}[thp]
  \vspace{-1.5em}
  \centering
  \caption{Corpus statistics. The $\tau^2$ column is the external, eval-only
  rendering of $\tau^2$-bench tasks.}
  \label{tab:data}
  \footnotesize
  \setlength{\tabcolsep}{4pt}
  \begin{tabular}{lcc}
    \toprule
    & Main corpus & $\tau^2$ rendering \\
    \midrule
    Dialogues & 26{,}694 & 274 \\
    Held-out dialogues (valid / test) & 210 / 210 & 0 / 274 \\
    Domains & 48 & 3 \\
    Audio & 786\,h & 4.9\,h \\
    \textsc{ready} / \textsc{unsure} events & 28{,}791 / 24{,}701 & 238 / 11 \\
    Test gold events & 463 & 249 \\
    Gold $\Delta_{\text{save}}$ med / p90 (s) & 1.36 / 6.60 & 0.00 / 3.87 \\
    \bottomrule
  \end{tabular}
  \vspace{-0.5em}
\end{table}

We assemble one sequence per dialogue. The system prompt and the profile's tool schemas are followed by a boundary token, the chunk's 8 audio embeddings $\mathbf{x}_c$ (0.6\,s of audio), and the emitted tokens $\mathbf{y}_c$ predicted as $p(\mathbf{y}_c \mid \mathbf{x}_{\le c}, \mathbf{y}_{<c})$. Every $\mathbf{y}_c$ begins with a VAD token $v_c\in\{\textsc{spk},\textsc{sil}\}$ and if $v_c{=}\textsc{spk}$ then an \etf{} state $s_c$. \textsc{ready} and \textsc{unsure} open a payload block, and three \emph{payload variants} differ only in what it contains: 
\begin{equation}
\label{eq:variants}
\mathbf{y}_c=\begin{cases}
(v_c,\,s_c) & \armnone{}\\
(v_c,\,s_c,\,a_c\,[s_c{\in}\{\textsc{r},\textsc{u}\}]) & \armasr{}\\
(v_c,\,s_c,\,a_c\,[s_c{\in}\{\textsc{r},\textsc{u}\}],\,j_c\,[s_c{=}\textsc{r}]) & \armaj{}
\end{cases}
\end{equation}
where $[\cdot]$ is the emission condition, \textsc{r}/\textsc{u} abbreviate \textsc{ready}/\textsc{unsure}, $a_c$ is a partial-ASR span of the speech heard so far between \texttt{<asr>} delimiters, and $j_c$ is the tool-call JSON (\texttt{\{"tool\_name", "arguments"\}}, keys sorted) between \texttt{<tool>} delimiters.
Each variant answers differently how a fire becomes an executable call: \armaj{} is self-contained, its \textsc{ready} JSON being the call; \armasr{} externalizes formatting, sending its span, the tool schemas, and the dialogue prefix to a frozen LLM formatter that emits the JSON; \armnone{} carries no payload, so acting on its fires also requires an external ASR pass over the audio near the fire.

\section{Experiments}
\label{sec:exp}

\subsection{Setup}
\label{sec:setup}

All variants are trained with LoRA rank $r{=}16$ (843.8M trainable parameters), under one weighted cross-entropy (VAD 0.4; \etf{}, ASR, JSON 1.0; \textsc{ready}/\textsc{unsure} class weights 6/10) with event-count-aware oversampling, early stopping, and a single seed. 
We decode the held-out test split and the audio-rendered $\tau^2$-bench in streaming mode, with a finite-state machine masking illegal tokens per slot, and score at two levels of granularity.
At the \emph{state} level, every chunk's VAD state and every speech chunk's \etf{} state are classified against gold with exact alignment. At the \emph{event} level, each \textsc{ready}/\textsc{unsure} emission is a predicted event and is \emph{matched} if a gold event of the same state lies within $\pm 2$ chunks ($\pm1.2$\,s) regardless of payload.
This rule yields the event metrics of Table~\ref{tab:main} and the \emph{matched fires} that form the denominator of Table~\ref{tab:cascade} and Fig.~\ref{fig:headroom}.
The tolerance reflects the coarse 0.6\,s grid: a gold $t_{\text{fire}}$ near a chunk boundary is legitimately answered from the neighboring chunk.
We report event recall/precision/F1, \etf{}/VAD-state macro-F1, per-state F1, and word error rate (WER) of emitted partial-ASR spans; at the call level, we report on matched \textsc{ready} fires, exact match (EM) of the JSON string, tool-name accuracy, and argument accuracy (micro-averaged over gold slots). 
The formatter is Qwen3-235B-A22B or Qwen3-30B-A3B \cite{qwen3}, prompted with the tool pool, the completed turns before the event, previous calls, and the model's emissions.

The \emph{achieved saving} is $t_{\text{end}} - t^{\text{pred}}_{\text{fire}}$, charging a predicted fire at the end of its chunk. It estimates the savings over the turn-end cascade \armturn{}, which transcribes the completed turn and hands it to the same formatter, which also decides whether to call.
\armturnasr{} uses Parakeet-TDT-CTC 1.1B \cite{fastconformer,tdt}, the aligner of our data pipeline, as external ASR; \armturngold{} uses the gold transcript; \armnone{} runs the same ASR over the audio window around its fire. The baseline transcript is charged at zero latency, an ideal streaming recognizer, and the assumption most favorable to the baseline.
\subsection{Results}
\label{sec:results}

\begin{table}[t]
  \centering
  \caption{Decision quality on the in-domain (ID) and audio-rendered $\tau^2$-bench splits for event, state, and WER across three variants. Bold marks the best variant per row within each split.
  }
  \label{tab:main}
  \footnotesize
  \setlength{\tabcolsep}{2.2pt}
  \begin{tabular}{lcccccc}
    \toprule
    Payload variant & \multicolumn{2}{c}{\armnone{}} & \multicolumn{2}{c}{\armasr{}}
    & \multicolumn{2}{c}{\armaj{}} \\
    \cmidrule(lr){2-3}\cmidrule(lr){4-5}\cmidrule(lr){6-7}
    Test split & ID & $\tau^2$ & ID & $\tau^2$ & ID & $\tau^2$ \\
    \midrule
    Event recall & \textbf{0.553} & 0.558 & 0.510 & 0.562 & 0.495 & \textbf{0.582} \\
    Event precision & 0.123 & 0.147 & \textbf{0.349} & \textbf{0.412} & 0.329 & 0.409 \\
    Event F1 & 0.201 & 0.232 & \textbf{0.414} & 0.475 & 0.395 & \textbf{0.480} \\
    \etf{}-state macro-F1 & 0.477 & 0.425 & \textbf{0.484} & \textbf{0.474} & 0.463 & 0.461 \\
    VAD-state macro-F1 & 0.993 & 0.976 & \textbf{0.996} & \textbf{0.982} & 0.994 & 0.979 \\
    Partial-ASR span WER & --- & --- & 0.120 & 0.551 & \textbf{0.112} & \textbf{0.478} \\
    \bottomrule
  \end{tabular}
  \vspace{-2em}
\end{table}

\begin{table}[t]
  \centering
  \caption{Per-class \etf{} F1, computed per chunk over all speech chunks for the in-domain test split. F1@$+k$ credits a prediction up to $k$ chunks ($0.6k$\,s) late via exact matching; @0 is exact alignment. Bold marks the best variant per column at @0.}
  \label{tab:perclass}
  \footnotesize
  \setlength{\tabcolsep}{3.6pt}
  \begin{tabular}{llccccc}
    \toprule
    & & \textsc{wait} & \textsc{ready} & \textsc{unsure} & \textsc{hold} & Macro \\
    \midrule
    \armnone{} & @0 & \textbf{0.936} & 0.309 & 0.175 & \textbf{0.488} & 0.477 \\
     & @$+1$ & 0.938 & 0.405 & 0.242 & 0.493 & 0.519 \\
     & @$+3$ & 0.938 & 0.432 & 0.272 & 0.493 & 0.534 \\
    \midrule
    \armasr{} & @0 & 0.932 & \textbf{0.315} & \textbf{0.206} & 0.483 & \textbf{0.484} \\
     & @$+1$ & 0.934 & 0.384 & 0.244 & 0.487 & 0.512 \\
     & @$+3$ & 0.934 & 0.406 & 0.277 & 0.487 & 0.526 \\
    \midrule
    \armaj{} & @0 & 0.927 & 0.299 & 0.158 & 0.469 & 0.463 \\
     & @$+1$ & 0.929 & 0.359 & 0.205 & 0.471 & 0.491 \\
     & @$+3$ & 0.929 & 0.367 & 0.232 & 0.471 & 0.500 \\
    \bottomrule
  \end{tabular}
  \vspace{-1.5em}
\end{table}

Table~\ref{tab:main} reports the decision layer on both domains. All three variants detect events with recall around 0.5 on ID and 0.56--0.58 on $\tau^2$, though emitting a payload does sharpen the decision: the payload-free \armnone{} fires far too often, with precision of only 0.123 and 0.147 for ID and $\tau^2$, while \armasr{}, which transcribes at the fire, gives the best ID event F1 of 0.414 and the best state classification for both \etf{} and VAD; the \armasr{}/\armaj{} differences are small.
The decision layer carries over to $\tau^2$ largely intact: \etf{}-state macro-F1 is unchanged for the span variants (only \armnone{} drops by 0.05), and event F1 comes out even higher, with precision of 0.412.
The payload is where transfer breaks: partial-ASR WER is 0.11--0.12 in domain but rises to 0.48 and 0.55 for \armaj{} and \armasr{} on $\tau^2$'s unseen entity-value patterns.

Table~\ref{tab:perclass} splits the \etf{}-state macro-F1 by state and by lateness tolerance. \textsc{wait} is nearly solved and \textsc{ready}/\textsc{unsure} carry the errors: false-positive \textsc{ready} fires are the precision bottleneck, and every variant learns \textsc{unsure} where \armasr{} gets best F1 of 0.206. Granting one chunk of lateness tolerance lifts \textsc{ready} F1 by 0.06--0.10 and the macro by 0.03--0.04, with little further gain at $+3$: a sizeable share of the apparent errors are fires one chunk late.

\begin{table}[t]
  \centering
  \caption{Firing decision under one turn-level rule for in-domain test split (1{,}499 user turns, 292 with a gold \textsc{ready} event). A system fires on a turn if it emits at least one \textsc{ready} in it; a fire is a true positive if that turn holds a gold \textsc{ready}. The payload variants fire from their own streaming \textsc{ready} decisions; \armturn{} fires when the formatter decides to call. Bold marks the best per column.}
  \label{tab:turnfire}
  \footnotesize
  \setlength{\tabcolsep}{4pt}
  \begin{tabular}{llcccc}
    \toprule
    Variant & Decision from & Fired & P & R & F1 \\
    \midrule
    \armasr{} & model \textsc{ready} & 458 & 0.376 & 0.589 & 0.459 \\
    \armaj{} & model \textsc{ready} & 418 & \textbf{0.409} & 0.586 & 0.482 \\
    \armturnasr{} & Parakeet text & 789 & 0.331 & \textbf{0.894} & 0.483 \\
    \armturngold{} & gold text & 751 & 0.348 & \textbf{0.894} & \textbf{0.500} \\
    \bottomrule
  \end{tabular}
\end{table}

Table~\ref{tab:turnfire} answers the question: \textbf{on a turn that needs a call, did the system fire?} It scores every system under one turn-level rule, so our systems can be set against the turn-end cascade \armturn{}.
\armturn{} is recall-heavy; at recall 0.894 it fires on 751--789 of the 1{,}499 turns, two-thirds of them without a gold \textsc{ready}, whereas \armaj{} gives up 0.3 of recall for the highest precision of 0.409; F1 is within 0.05 across all four.
Deciding early thus trades off recall for precision at the same F1. Even so, 59\% of early fires are spurious at this operating point, so the precision an agent needs before committing a call has not yet been reached.

\begin{figure}[t]
  \centering
  \includegraphics[width=\columnwidth]{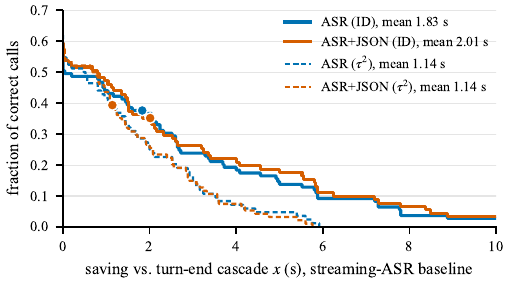}
  \caption{Achieved saving of \emph{correct} early calls (Qwen3-235B-A22B formatter) over the \armturn{} baseline: fraction of calls saving more than $x$ seconds. Dots mark means (legend); the value at $x{=}0$ is the fraction that fires before the turn ends; medians over all correct calls are 0.0\,s (\armasr{}) and 0.7\,s (\armaj{}) on ID. }
  \label{fig:headroom}
\end{figure}

Figure~\ref{fig:headroom} shows the distribution of the achieved saving restricted to \emph{correct} calls, with the Qwen3-235B-A22B formatter, for both variants and both domains.
The distributions are heavily right-skewed. For ID, about half of the correct calls fire before the turn ends (50\% for \armasr{}, 54\% for \armaj{}); the rest fire in the turn's final chunk. The median saving over all correct calls is therefore 0.0\,s (\armasr{}) and 0.7\,s (\armaj{}); among the calls that do fire early, the median is 2.6\,s, and 44\% and 47\% of correct calls save more than 1\,s, so the means (1.83 and 2.01\,s, legend) are carried by a long tail.
On $\tau^2$, gold $\Delta_{\text{save}}$ has a median of 0 because its turns place tool evidence at the turn end, which makes the timing task easier; even so, 54--55\% of the correct calls fire before the turn end, with a median saving of 1.8\,s among them.

\begin{table}[hpt]
  \centering
  \caption{Call quality by payload variant. \emph{Variant}: whose
  \textsc{ready} fires trigger the call. \emph{Call from}: the text the call is built from, with the model's own JSON (no formatter), the model's own span ($a_c$), a Parakeet transcript, or the gold transcript; the latter three are via the formatter. For \armturn{} the text covers the completed turn; for all other variants, the audio up to the fire. EM, tool-name, and argument accuracy on matched fires (\armturn{}: on true-positive turns). Bold marks the best per column (except GT ASR).}
  \label{tab:cascade}
  \footnotesize
  \setlength{\tabcolsep}{1.5pt}
  \begin{tabular}{lllcccccc}
    \toprule
    & & & \multicolumn{3}{c}{ID} & \multicolumn{3}{c}{$\tau^2$} \\
    Variant & Call from & Formatter & EM & Tool & Arg & EM & Tool & Arg \\
    \midrule
    \armaj{} & model JSON & --- & 0.173 & 0.232 & 0.519 & 0.000 & 0.073 & 0.054 \\
    \armaj{} & model ASR & 235B & 0.536 & 0.833 & 0.710 & \textbf{0.681} & \textbf{0.978} & \textbf{0.844} \\
    \armaj{} & model ASR & 30B & 0.530 & 0.833 & 0.707 & 0.674 & 0.964 & 0.839 \\
    \armasr{} & model ASR & 235B & \textbf{0.578} & 0.840 & \textbf{0.711} & 0.618 & 0.971 & 0.770 \\
    \armasr{} & model ASR & 30B & 0.535 & 0.845 & 0.683 & 0.588 & 0.963 & 0.776 \\
    \armasr{} & GT ASR & 235B & 0.626 & 0.840 & 0.760 & 0.838 & 0.963 & 0.888 \\
    \armasr{} & GT ASR & 30B & 0.588 & 0.856 & 0.737 & 0.831 & 0.963 & 0.893 \\
    \armnone{} & ext.\ ASR & 235B & 0.038 & \textbf{0.870} & 0.493 & 0.090 & 0.955 & 0.696 \\
    \midrule
    \armturn{} & ext.\ ASR & 235B & 0.103 & 0.789 & 0.634 & 0.073 & 0.744 & 0.753 \\
    \armturn{} & GT ASR & 235B & 0.605 & 0.751 & 0.832 & 0.558 & 0.694 & 0.915 \\
    \bottomrule
  \end{tabular}
\end{table}

\section{Ablation: producing the call}
\label{sec:ablation}

Table~\ref{tab:cascade} compares the strategies on the models' own \textsc{ready} fires.

\emph{Externalizing the call is clearly better.} On \armaj{}'s own fires, the formatter given the model's span reaches EM 0.536, against 0.173 for the JSON the model emits itself, and the decomposition shows why: the trained JSON head names the tool on only 23\% of matched fires and fills 52\% of argument slots, whereas the formatter reaches 0.83 and 0.71 from the same span. The \armasr{} variant, trained without a JSON head, still performs slightly better with EM of 0.578.

\emph{Span fidelity, not the formatter, is the residual.} The gold-text rows bound the remaining loss: tool naming is unchanged with ACC of 0.84 while argument accuracy rises to 0.76 and EM to 0.63, so what the model transcribes at the fire moment is the dominant error; the 30B formatter concedes at most 4.3 EM points.

\emph{A bare decision is not enough.} The decision-only \armnone{} variant names the tool from a Parakeet pass over the fire window with Tool ACC of 0.87, but fills only half the argument slots, and its EM falls to 0.038 because Parakeet output carries spoken-form entities that never string-match the gold JSON.
The model's own transcribe-at-fire span is what makes the call executable (on $\tau^2$, the same variant improves to EM 0.090 on shorter, more regular payloads, still roughly $7\times$ below the span variants).

\emph{Against the turn-end cascade.} The \armturn{} rows are the deployed alternative. With Parakeet text, \armturnasr{} names the tool on 79\% of its true-positive turns and fills 63\% of argument slots, against 83\% of slots with gold text, yet scores EM 0.103: the same normalization failure, not timing. 
With gold text, it reaches EM 0.605. The pattern transfers: on $\tau^2$, the Parakeet arm again falls to EM 0.073 while gold text holds 0.558.

\section{Conclusion}
\label{sec:concl}

This paper presented spoken tool calling as an earliest-actionable-point decision. A synthetic pipeline labels, for 26,694 dialogues, the instant each call's tool and arguments are grounded in the audio, with disjoint train/test values; a LoRA-adapted Qwen3-Omni-30B-A3B decides per 0.6\,s chunk whether that instant has arrived and emits the call as JSON or a partial transcript for an off-the-shelf formatter. Under streaming decoding on held-out dialogues and audio-rendered $\tau^2$-bench, the decision reaches macro-F1 of 0.48 and 0.47, and correct early calls start an estimated 1.8--2.0\,s sooner than a turn-end cascade with the same formatter. The transcript yields exact calls $3\times$ as often as self-emitted JSON (0.54 vs.\ 0.17 on the same fires), and a payload-free decision plus external ASR fills half the argument slots. The evidence is single-seed on one base model; with 59\% of early fires spurious, fire precision is the open gap.

\vfill\pagebreak

\bibliographystyle{IEEEbib}
\bibliography{refs}

\end{document}